\documentclass{article}

\usepackage{microtype}
\usepackage{graphicx}
\usepackage{subfigure}
\usepackage{booktabs} 
\usepackage{amsmath}
\DeclareMathOperator*{\argmax}{arg\,max}

\usepackage{hyperref}

\usepackage[accepted]{icml2021}

\icmltitlerunning{Temporal Predictive Coding For Model-Based Planning In Latent Space}
\usepackage{mycustom}
\usepackage{xspace}
\newcommand{\RCPC}{TPC\xspace}

\newcommand{\sharedtitle}{Temporal Predictive Coding For Model-Based Planning In Latent Space}

\begin{document}

\twocolumn[
\icmltitle{\sharedtitle}



\icmlsetsymbol{equal}{*}

\begin{icmlauthorlist}
\icmlauthor{Tung Nguyen}{equal,vinai}
\icmlauthor{Rui Shu}{equal,stanford}
\icmlauthor{Tuan Pham}{equal,vinai}
\icmlauthor{Hung Bui}{vinai}
\icmlauthor{Stefano Ermon}{stanford}
\end{icmlauthorlist}

\icmlaffiliation{vinai}{VinAI Research}
\icmlaffiliation{stanford}{Stanford University}

\icmlcorrespondingauthor{Tung Nguyen}{v.tungnd13@vinai.io}
\icmlcorrespondingauthor{Rui Shu}{ruishu@stanford.edu}

\icmlkeywords{Machine Learning, ICML}

\vskip 0.3in
]



\printAffiliationsAndNotice{\icmlEqualContribution} 

\begin{abstract}
High-dimensional observations are a major challenge in the application of model-based reinforcement learning (MBRL) to real-world environments. To handle high-dimensional sensory inputs, existing approaches use representation learning to map high-dimensional observations into a lower-dimensional latent space that is more amenable to dynamics estimation and planning. 
In this work, we present an information-theoretic approach that employs \emph{temporal} predictive coding to encode elements in the environment that can be predicted across time.
Since this approach focuses on encoding temporally-predictable information, we implicitly prioritize the encoding of task-relevant components over nuisance information within the environment that are provably task-irrelevant. 
By learning this representation in conjunction with a recurrent state space model, we can then perform planning in latent space. 
We evaluate our model on a challenging modification of standard DMControl tasks where the background is replaced with natural videos that contain complex but irrelevant information to the planning task. Our experiments show that our model is superior to existing methods in the challenging complex-background setting while remaining competitive with current state-of-the-art models in the standard setting.

\end{abstract}

\section{Introduction}

Learning to control from high dimensional observations has been made possible due to the advancements in reinforcement learning (RL) and deep learning. These advancements have enabled notable successes such as solving video games \citep{mnih2015human, 10.5555/3298483.3298548} and continuous control problems \citep{DBLP:journals/corr/LillicrapHPHETS15} from pixels. However, it is well known that performing RL directly in the high-dimensional observation space is sample-inefficient and may require a large amount of training data \citep{lake2017building}. This is a critical problem, especially for real-world applications. Recent model-based RL works \citep{Kaiser2020Model,NIPS2018_7512,pmlr-v97-hafner19a,pmlr-v97-zhang19m,Hafner2020Dream} proposed to tackle this problem by learning a world model in the latent space, and then applying RL algorithms in the latent world model.


The existing MBRL methods that learn a latent world model typically do so via reconstruction-based objectives, which are likely to encode task-irrelevant information, such as of the background. In this work, we take inspiration from the success of contrastive learning and propose a model that employs temporal predictive coding for planning from pixels. 
The use of temporal predictive coding circumvents the need for reconstruction-based objectives and prioritizes the encoding of temporally-predictable components of the environment, thus making our model robust to environments dominated by nuisance information.
Our primary contributions are as follows:
\begin{itemize}
    \item We propose a temporal predictive coding approach for planning from high-dimensional observations and theoretically analyze its ability to prioritize the encoding of task-relevant information.
    \item We show experimentally that temporal predicting coding allows our model to outperform existing state-of-the-art models when dealing with complex environments dominated by task-irrelevant information, while remaining competitive on standard DeepMind control (DMC) tasks. Additionally, we conduct detailed ablation analyses to characterize the behavior of our model.
\end{itemize}

\section{Motivation}

The motivation and design of our model are largely based on two previous works \citep{shu2020predictive,Hafner2020Dream}. In this section, we briefly go over the relevant concepts in each work as well as how they motivate our paper. 
\citet{shu2020predictive} proposed PC3, 
an information-theoretic approach that uses contrastive predictive coding (CPC) to learn a latent space amenable to locally-linear control. Their framework uses a CPC objective between the latent states of two consecutive time steps. They then use the latent dynamics $F$ as the variational device to define the following lower bound $\ell_{\text{cpc}}(E,F)$:
\begin{equation}\label{eq:pc3_cpc}
    \Expect \frac{1}{K} \sum_{i} \ln \frac{F\left(E\left(o_{t+1}^{(i)}\right) \mid E\left(o_{t}^{(i)}\right), a_{t}^{(i)}\right)}{\frac{1}{K} \sum_{j} F\left(E\left(o_{t+1}^{(i)}\right) \mid E\left(o_{t}^{(j)}\right), a_{t}^{(j)}\right)} ,
\end{equation}
in which $E$ is the encoder, which encodes a high-dimensional observation into a latent state. We make particular note of their choice to define the CPC objective for latent states \emph{across} time steps instead of between the frame and its corresponding state---as is popular in the existing literature on contrastive learning for RL \cite{Hafner2020Dream,srinivas2020curl,ding2020mutual}. We shall refer to these respective objectives henceforth as \emph{temporal} predictive coding (TPC) and \emph{static} predictive coding (SPC) respectively. \citet{shu2020predictive} motivated temporal predictive coding via a theory of predictive suboptimality, which shows that the representation learned by temporal predictive coding is no worse at future-observation prediction than the representation learned by a latent variable model when applied to the same task. 

In this work, we take a stronger position and argue that temporal predictive coding learns an information-theoretically \emph{superior} representation to conventional latent variable models and static predictive coding models. This is because conventional latent variables models (i.e. temporal variational autoencoders equipped with a Gaussian decoder) and static predictive coding models both seek to encode as much information as possible about the original high-dimensional observations. In contrast, temporal predictive coding \emph{only} encourages the encoding of temporally-predictable information within the environment, which we shall prove is sufficient for optimal decision making (\cref{sec:theory}). This property of temporal predictive coding thus makes this objective particularly suitable for handling RL problems dominated by nuisance information, which has recently become an active area of research \cite{ding2020mutual,zhang2020learning,ma2020contrastive}.

However, since PC3 only tackles the problem from an optimal control perspective, it is not readily applicable to RL problems. Indeed, PC3 requires a depiction of the goal to perform control, and also the ability to teleport to random locations of the state space to collect data, which are impractical in many problems. On the other hand, Dreamer \citep{Hafner2020Dream} achieves state-of-the-art performance on many RL tasks, but learns the latent space using a reconstruction-based objective.
And while the authors reported a contrastive approach that yielded inferior performance to their reconstruction-based approach, their contrastive approach employed a static predictive coding objective. 

The goal of our paper is thus two-fold. First, we propose a model that leverages the concepts in PC3 and apply them to the Dreamer paradigm and RL setting. In particular, we show how to incorporate temporal predictive coding into the training of the recurrent state space model used in Dreamer. Second, we will demonstrate both theoretically and empirically that our resulting model is robust to RL environments that are dominated by nuisance information.

\section{Temporal Predictive Coding for Planning}
To plan in an unknown environment, we need to model the environment dynamics from experience. Following the Dreamer paradigm, we do so by iteratively collecting new data and using those data to train the world model. In this section, we focus on presenting the proposed model (its components and objective functions), a theoretical analysis of temporal predictive coding, followed lastly by some practical considerations when implementing the method.

\subsection{Model Definition}\label{sec:model}
\begin{figure*}[ht]
\centering
\begin{tikzpicture}[scale=0.8, every node/.style={scale=0.7}]
\node[circ, fill=gray] (o0) {$o_0$};
\node[circ, xshift=0cm, yshift=-2.5cm] (s0) {$s_0$};
\node[box, xshift=1.5cm, yshift=-5cm] (h1) {$h_1$};
\node[circ, xshift=3cm, yshift=-2.5cm] (shat1) {$\hat{s}_1$};

\node[none, xshift=4cm, yshift=-2.5cm] (approx1) {$\approx$};

\node[circ, fill=gray, xshift=5cm, yshift=0cm] (o1) {$o_1$};
\node[circ, xshift=5cm, yshift=-2.5cm] (s1) {$s_1$};
\node[box, xshift=6.5cm, yshift=-5cm] (h2) {$h_2$};
\node[circ, xshift=8cm, yshift=-2.5cm] (shat2) {$\hat{s}_2$};

\node[none, xshift=9cm, yshift=-2.5cm] (approx2) {$\approx$};

\node[circ, fill=gray, xshift=10cm, yshift=0cm] (o2) {$o_2$};
\node[circ, xshift=10cm, yshift=-2.5cm] (s2) {$s_2$};
\node[box, xshift=11.5cm, yshift=-5cm] (h3) {$h_3$};
\node[circ, xshift=13cm, yshift=-2.5cm] (shat3) {$\hat{s}_3$};


\path
(h1) edge [connect] (h2)
(h2) edge [connect] (h3)
(o0) edge [connect] (s0)
(o1) edge [connect] (s1)
(o2) edge [connect] (s2)
(h1) edge [connect] (shat1)
(h2) edge [connect] (shat2)
(h3) edge [connect] (shat3)
(s0) edge [connect] (h1)
(s1) edge [connect] (h2)
(s2) edge [connect] (h3)

;
\end{tikzpicture}  
\caption{The graphical model of TPC, in which we employ a recurrent neural network to model the dynamics. We omit the action and reward for simplicity. The approximation of $\hat{s}_t$ with $s_t$ is done by using contrastive predictive coding and maximum likelihood.}
\label{fig:model}
\end{figure*}
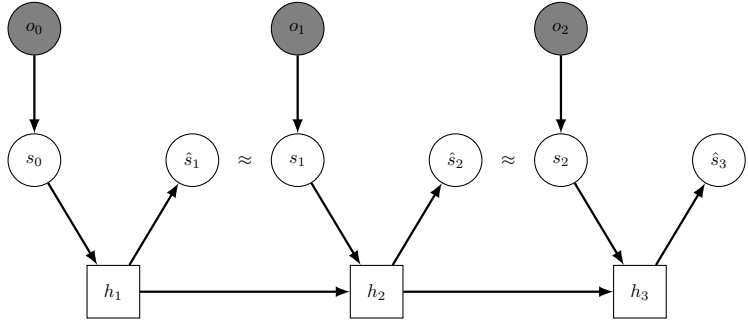

Since we aim to learn a latent dynamics model for planning, we shall define an encoder $E$ to embed high-dimensional observations into a latent space, a latent dynamics $F$ to model the world in this space, and a reward function,
\begin{equation}
    \begin{aligned}
        & \text{Encoder:} && E(o_t) = s_t \\
        & \text{Latent dynamics:} && F(s_t \giv s_{<t}, a_{<t}) = p(s_t \giv s_{<t}, a_{<t}) 
        \\
        & \text{Reward function:} && R(r_t \giv s_t) = p(r_t \giv s_t)
    \end{aligned}
\end{equation}
in which $t$ is the discrete time step, $\{o_t,a_t,r_t\}_{t=1}^T$ are data sequences with image observations $o_t$, continuous action vectors $a_t$,  scalar rewards $r_t$, and $s_t$ denotes the latent state at time $t$. We model the transition dynamics using a recurrent neural network with a deterministic state $h_t = \text{RNN}(h_{t-1}, s_{t-1}, a_{t-1})$, which summarizes information about the past, followed by the stochastic state model $p(s_t | s_{<t}, a_{<t}) = p(s_t | h_t)$. Following \citet{pmlr-v97-hafner19a}, we refer to this RNN as the recurrent state space model (RSSM). In practice, we use a deterministic encoder, and Gaussian distribution for dynamics and reward functions. The graphical model is presented in Figure \ref{fig:model}. We now introduce the three key components of our model: \textit{temporal predictive coding}, \textit{consistency}, and \textit{reward prediction}. 

\paragraph{Temporal predictive coding with an RSSM} Instead of performing pixel prediction to learn $E$ and $F$ as in \citet{pmlr-v97-hafner19a,Hafner2020Dream}, we maximize the mutual information between the past latent codes and actions against the future latent code $I(E(O_{t}); E(O_{<t}), A_{<t})$. This objective prioritizes the encoding of predictable components from the environment, which potentially helps avoid encoding nuisance information when dealing with complex image observations. Unlike in \citet{shu2020predictive}, which defined the TPC objective in a Markovian setting, the use of an RSSM means that we are maximizing the mutual information between the latent code at any time step $t$ and the entire historical trajectory latent codes and actions prior to $t$. To handle a trajectory over $T$ steps, we define an objective that sums over every possible choice of $t \in \{2, \ldots, T\}$,
\begin{equation}
    \sum_{t=2}^T I(E(O_t); E(O_{<t}), A_{<t}).
\end{equation}

To estimate this quantity, we employ contrastive predictive coding (CPC) proposed by \citet{oord2018representation}. We perform CPC by introducing a critic function $f$
to construct the lower bound at a particular time step $t$,
\begin{align}\label{eq:mi}
    & I(E(O_{t}); E(O_{<t}), A_{<t}) \geq \\
    & \Expect \frac{1}{K} \sum_i \ln \frac{\exp f(E(o_t^{(i)}), E(o_{<t}^{(i)}), a_{<t}^{(i)})}{\frac{1}{K} \sum_j \exp f(E(o_t^{(i)}), E(o_{<t}^{(j)}), a_{<t}^{(j)})} =: \ell^{(t)}_\text{tpc}, \nonumber
\end{align}
where the expectation is over $K$ i.i.d. samples of $(o_t, o_{<t}, a_{<t})$. Note that \cref{eq:mi} uses past $(E(o^{(j)}_{<t}), a^{(j)}_{<t})$ from unrelated trajectories as an efficient source of negative samples for the contrastive prediction of the future latent code $E(o^{(i)}_t)$. Following \citet{shu2020predictive}, we choose to tie $f$ to our recurrent latent dynamics model $F$,
\begin{equation}
    \exp f(s_t, s_{<t}, a_{<t}) = F(s_t | s_{<t}, a_{<t}).
\end{equation}
There are two favorable properties of this particular design. First, it is parameter-efficient since we can circumvent the instantiation of a separate critic $f$. Moreover, it takes advantage of the fact that an optimal critic is the true latent dynamics induced by the encoder $E$ \citep{pmlr-v97-poole19a,ma-collins-2018-noise}. We denote our objective as $\ell_\text{tpc}(E, F) = \sum_{t=2}^T\ell^{(t)}_\text{tpc}(E, F)$.

\paragraph{Enforcing latent dynamics consistency} 
Although the true dynamics is an optimal critic for the CPC bound, maximizing this objective only does not ensure the learning of a latent dynamics model $F$ that is consistent with the true latent dynamics, due to the non-uniqueness of the optimal critic \citep{shu2020predictive}. Since an accurate dynamics is crucial for planning in the latent space, we additionally introduce a consistency objective, which encourages the latent dynamics model to maintain a good prediction of the future latent code given the past latent codes and actions. Similar to the recurrent CPC objective, we optimize for consistency at every time step in the trajectory,
\begin{equation}
    \ell_{\text{cons}}(E,F) = \sum_{t=2}^T \Expect_{p(o_t, o_{<t}, a_{<t})} \ln F(E(o_t) | E(o_{<t}), a_{<t}).
\end{equation}

\paragraph{Reward prediction} Finally, we train the reward function by maximizing the likelihood of the true reward value conditioned on the latent state $s_t = E(o_t)$,
\begin{equation}\label{eq:reward}
    \ell_{\text{reward}}(E,R) = \sum_{t=1}^T \Expect_{p(o_t)} \ln R(r_t | E(o_t)).
\end{equation}

\subsection{Theoretical Analysis}\label{sec:theory}


In contrast to a reconstruction-based objective, which explicitly encourages the encoder to behave injectively on the space of observations, our choice of mutual information objective as specified in \cref{eq:mi} may discard information from the observed scene. In this section, we wish to formally characterize the information discarded by the temporal predictive coding (TPC) objective and argue that any information discarded by an optimal encoder under the TPC objective is provably task-irrelevant. 

\begin{lemma}
\label{lem:relevance}
Consider an optimal encoder and reward predictor pair $(E^*, R^*)$ where
\begin{align}
    \argmax_E I(E(O_t) \scolon E(O_\ltt), A_\ltt) &= E^*\\
    \kldiv{p(r_t \giv o_t)}{R^*(r_t \giv E^*(o_t))} &= 0.
\end{align}
Let $\pi(a_t \giv E^*(o_\leqt), a_\ltt)$ denote an $E^*$-restricted policy whose access to the observations $o_\ltt$ is restricted by $E^*$. Let $\pi_\aux(a_t \giv E^*(o_\leqt), E'(o_\leqt), a_\ltt)$ denote an $(E^*, E')$-restricted policy which has access to auxiliary information about $o_\ltt$ via some encoder $E'$. Let $\eta(\pi)$ denote the expected cumulative reward achieved by a policy $\pi$ over a finite horizon $T$. Then there exists no encoder $E'$ where the optimal $E^*$-restricted policy underperforms the optimal $(E^*, E')$-restricted policy,
\begin{align}
    \nexists E' \textrm{ s.t. } \eta(\pi^*) < \eta(\pi^*_\aux).
\end{align}
\end{lemma}

Intuitively, since $E^*$ optimizes TPC, any excess information contained in $E'$ about $o_t$ (not already accounted for by $E^*$) must be \emph{temporally-unpredictive}---it is neither predictable from the past nor predictive of the future. The excess information conveyed by $E'$ is effectively nuisance information and thus cannot be exploited to improve the agent's performance. It is therefore permissible to dismiss the excess information in $E'$ as being task-irrelevant. We provide a proof formalizing this intuition in \cref{app:lemma}.

It is worth noting that TPC does not actively penalize the encoding of nuisance information, nor does temporally-predictive information necessarily mean it will be task-relevant. However, if the representation space has limited capacity to encode information about $o_t$ (e.g., due to dimensionality reduction, a stochastic encoder, or an explicit information bottleneck regularizer), TPC will favor temporally-predictive information over temporally-unpredictive information---and in this sense, thus favor potentially task-relevant information over provably task-irrelevant information. This is in sharp contrast to a reconstruction objective, which makes no distinction between these two categories of information contained in the observation $o_t$.




\subsection{Practical Implementation Details}\label{sec:practical}

\paragraph{Avoiding map collapse} Naively optimizing $\ell_{\text{tpc}}$ and $\ell_{\text{cons}}$ can lead to a trivial solution, where the latent map collapses into a single point to increase the consistency objective arbitrarily. In the previous work, \citet{shu2020predictive} resolved this by adding Gaussian noise to the future encoding, which balances the latent space retraction encouraged by $\ell_{\text{cons}}$ with the latent space expansion encouraged by $\ell_{\text{tpc}}$. However, we found this trick insufficient and thus further incorporate a variant of static predictive coding. Our static predictive coding defines a lower bound for $I(E(O_t) \scolon O_t)$ where we employ a Gaussian distribution $\Normal(E(O_t) \giv E(O_t), \sigma I)$ as our choice of variational critic for $f(E(O_t), O_t)$. Crucially, our variant keeps the variance $\sigma$ fixed so that the static predictive coding objective solely serves the role of encouraging map expansion.


\paragraph{Smoothing the dynamics model} Since our encoder is deterministic, the dynamics always receives clean latent codes as inputs during training. However, in behavior learning, we roll out multiple steps towards the future from a stochastic dynamics. These roll outs are susceptible to a cascading error problem,
which hurts the value estimation and policy learning. To resolve this issue, we smooth the dynamics by adding Gaussian noise to the inputs of the recurrent dynamics during training. The noise-adding procedure is as follows: assume the dynamics outputs $\hat{s}_t \sim \Normal(\mu_t, \sigma^2_t)$ as the prediction at time step $t$, we then add $\epsilon_t \sim \Normal(0, \sigma^2_t)$ to $s_t = E(o_t)$ and feed it to the latent dynamics, and repeat for every time step $1 \leq t \leq T$. We call this dynamics-associated noise, which ensures that the latent dynamics can handle the amount of noise that it produces when rolling out. The overall objective of our model is
\begin{equation}
\begin{aligned}
    \max_{E,F, R} \text{ } & \lambda_1 \ell_{\text{tpc}}(E,F) + \lambda_2 \ell_{\text{cons}}(E,F) \\
    + & \lambda_3 \ell_{\text{spc}}(E) + \lambda_4 \ell_{\text{reward}}(E,R).
\end{aligned}
\end{equation}

\section{Behavior Learning}


Following \citet{Hafner2020Dream}, we use latent imagination to learn a parameterized policy for control. For self-containedness, this section gives a summary of this approach. Given the latent state $s_t = E(o_t)$, we roll out multiple steps into the future using the learned dynamics $F$, estimate the return and perform backpropagation through the dynamics to maximize this return, which in turn improves the policy.

\paragraph{Action and value models} Two components needed for behavior learning are the action model and the value model, which both operate on the latent space. The value model estimates the expected imagined return when following the action model from a particular state $s_\tau$, and the action model implements a policy, conditioned on $s_\tau$, that aims to maximize this return. With imagine horizon $H$, we have
\begin{equation}
    \begin{aligned}
        & \text{Action model:} &  & a_\tau \sim \pi(a_\tau|s_\tau)  \\
        & \text{Value model:} &  & v(s_\tau) \approx \Expect_{\pi(\cdot|s_\tau)} \sum_{\tau=t}^{t+H} \gamma^{\tau-t} r_\tau
    \end{aligned}
\end{equation}

\begin{figure*}[ht]
    \centering
    \includegraphics[width=.9\textwidth]{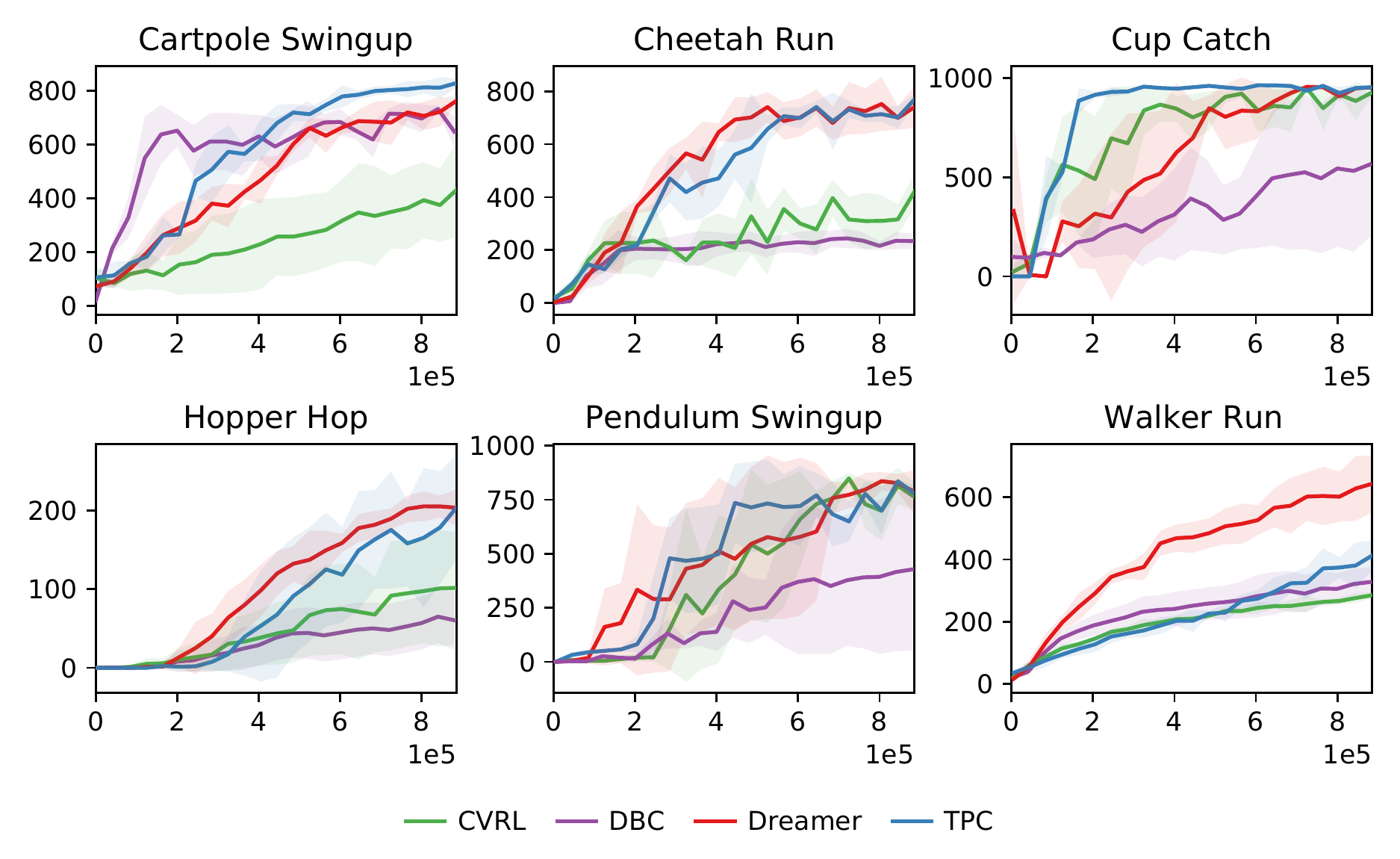}
    \vspace{-10pt}
    \caption{\RCPC vs the baselines in standard DMC tasks. The return is computed in 1000 environment steps. Each task is run with 3 seeds.}
    \label{fig:standard_compare}
\end{figure*}

\paragraph{Value estimation} To learn the action and value model, we need to estimate the state values of imagined trajectories $\{s_\tau, a_\tau, r_\tau \}_{\tau = t}^{t+H}$, where $s_\tau$ and $a_\tau$ are sampled according to the dynamics and the policy. In this work, we use value estimation presented in \citet{sutton2018reinforcement},
\begin{equation}
\begin{aligned}
\mathrm{V}_{\mathrm{N}}^{k}\left(s_{\tau}\right) & \doteq \mathrm{E}_{q_{\theta}, q_{\phi}}\left(\sum_{n=\tau}^{h-1} \gamma^{n-\tau} r_{n}+\gamma^{h-\tau} v_{\psi}\left(s_{h}\right)\right) \\ 
\mathrm{V}_{\lambda}\left(s_{\tau}\right) & \doteq (1-\lambda) \sum_{n=1}^{H-1} \lambda^{n-1} \mathrm{V}_{\mathrm{N}}^{n}\left(s_{\tau}\right)+\lambda^{H-1} \mathrm{V}_{\mathrm{N}}^{H}\left(s_{\tau}\right) \\
\text {with } h & =\min (\tau+k, t+H).
\end{aligned}
\end{equation}
which allows us to estimate the return beyond the imagine horizon. The value model is then optimized to regress this estimation, while the action model is trained to maximize the value estimation of all states $s_\tau$ along the imagined trajectories,

\begin{equation}
\begin{aligned}
& \max _{\pi} \Expect\left(\sum_{\tau=t}^{t+H} \mathrm{V}_{\lambda}\left(s_{\tau}\right)\right) \\
& \min _{v} \Expect \left(\sum_{\tau=t}^{t+H} \frac{1}{2} \| v_{\psi}\left(s_{\tau}\right)-\mathrm{V}_{\lambda}\left(s_{\tau}\right) \|^{2}\right).
\end{aligned}
\end{equation}

\begin{figure*}[ht]
    \centering
    \includegraphics[width=0.9\textwidth]{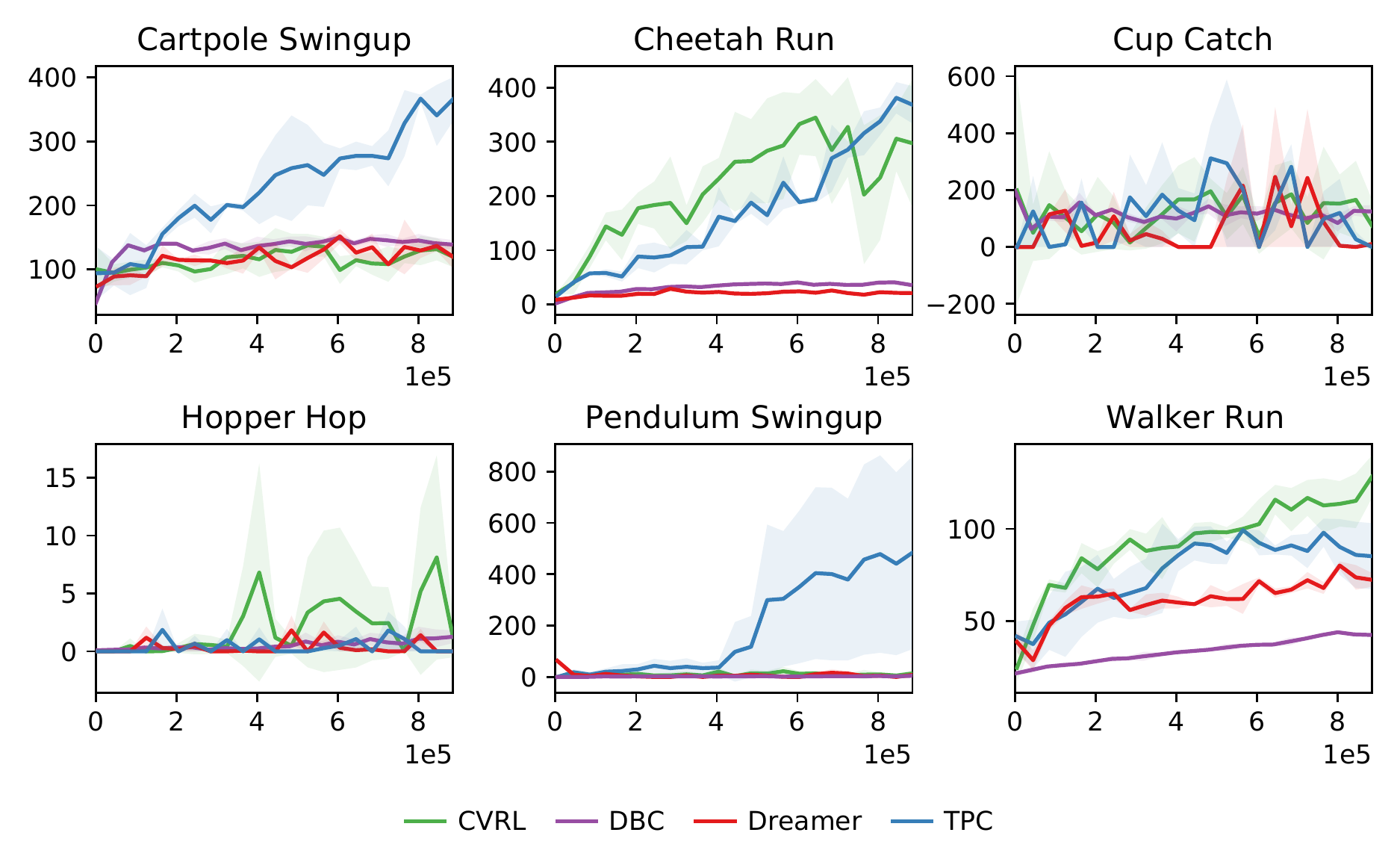}
    \vspace{-10pt}
    \caption{\RCPC vs the baselines in the natural backgrounds setting. The return is computed in 1000 evironment steps. Each task is run with 3 seeds.}
    \label{fig:background_compare}
\end{figure*}

\section{Experiments}
In this section, we empirically evaluate the proposed model (which we shall simply refer to as \RCPC for convenience) in various settings. First, we design experiments to compare the relative performance of our model with the current best model-based method in several standard control tasks. Second, we evaluate its ability to handle a more realistic but also more complicated scenario, in which we replace the background of the environment with a natural video. We discuss how our model is superior in the latter case compared to other existing methods, while remaining competitive in the standard setting. Finally, we conduct ablation studies to demonstrate the importance of the components of our model.

\paragraph{Control tasks} For the standard setting, we test our model on 6 DeepMind Control (DMC) tasks \citep{tassa2018deepmind}: Cartpole Swingup, Cheetah Run, Walker Run, Pendulum Swingup, Hopper Hop and Cup Catch. In the natural background setting, we replace the background of each data trajectory with a video taken from the kinetics dataset \citep{kay2017kinetics}. We split the original dataset into two separate sets for training and evaluation to also test the generalization of each method. 

\paragraph{Baseline methods} We compare \RCPC with Dreamer \citep{Hafner2020Dream}, CVRL \citep{ma2020contrastive}, and DBC \citep{zhang2020learning}. We compare against Dreamer since it is the current state of the art model-based method for planning from pixels and directly inspired our design of \RCPC. CVRL is a closely-related contrastive-based model recently developed that also touted its ability to handle nuisance backgrounds. Finally, we compare with DBC, a leading model-free method for learning representations that are invariant to distractors, to showcase how temporal predictive coding fares against bisimulation for learning task-relevant representations for downstream control. 
Each method is evaluated by the environment return in 1000 steps. For the baselines, we use the best set of hyperparameters as reported in their paper. We run each task with 3 different seeds for each model. 



\begin{figure*}[!ht]
    \centering
    \includegraphics[width=0.9\textwidth]{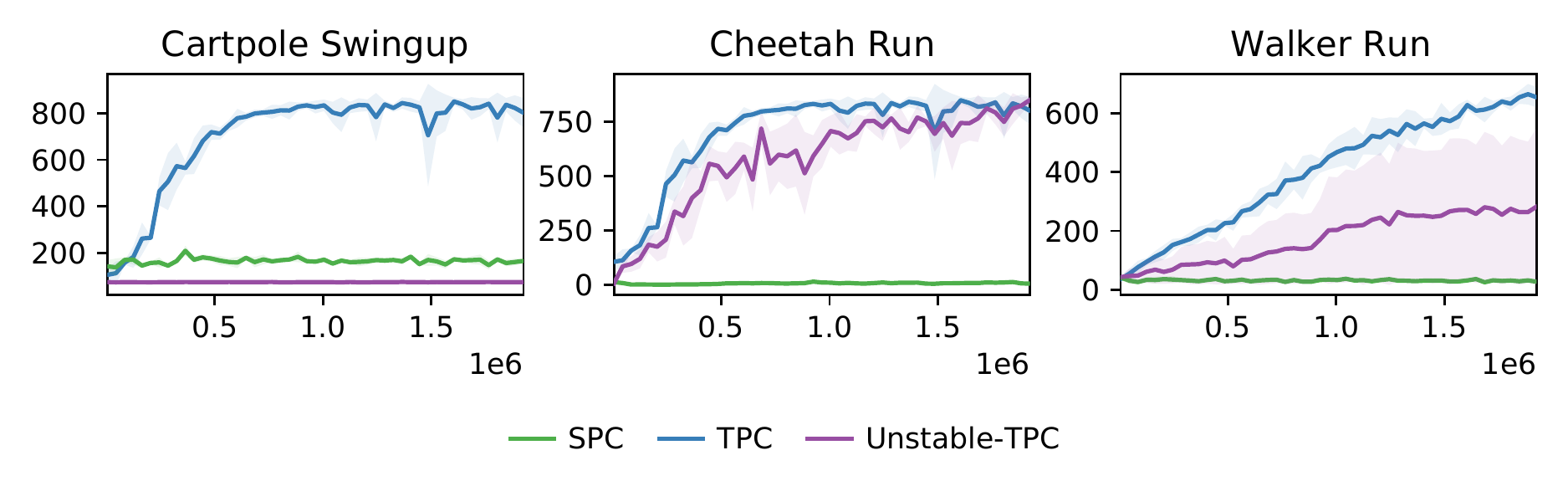}
    \vspace{-10pt}
    \caption{\RCPC vs two different variants, where we omit the temporal predictive coding in SPC and static predictive coding in Unstable-\RCPC. The return is computed in 1000 evironment steps. Each task is run with 3 seeds.}
    \label{fig:abaltion}
\end{figure*}

\subsection{Comparisons in Standard Setting}


We demonstrate the performance of all methods on standard control tasks in Figure \ref{fig:standard_compare}. \RCPC is competitive with Dreamer in all but one task, while CVRL and DBC both underperformed Dreamer on four tasks. \RCPC's strong performance is in contrast to what was previously observed in Dreamer \citep{Hafner2020Dream}, where they showed the inferiority of a \emph{static} contrastive learning approach (which applies CPC between an image observation and its corresponding latent code) compared to their reconstruction-based approach. Our results thus show that the use of \emph{temporal} contrastive predictive coding is critical for learning a latent space that is suitable for behavior learning.

\subsection{Comparisons in Natural Background Setting}

In this setting, we evaluate the robustness of \RCPC versus the baselines in dealing with complicated, natural backgrounds. 
The performance of all models in this setting is shown in \Cref{fig:background_compare}. Dreamer fails to achieve meaningful performance across all six tasks. This is due to the use of reconstruction loss, which forces the world model to reconstruct every single pixel in the observations, thus leading to the encoding of task-irrelevant information.

\RCPC outperforms all the baselines significantly on two of the six tasks, while being competitive to CVRL on Walker Run and Cheetah Run (see \cref{app:add_results} for more environment steps, where \RCPC outperforms CVRL on three tasks). All methods fail to work on Hopper Hop and Cup Catch\footnote{\RCPC and CVRL work on Cup Catch after $2 \times 10^6$ steps. The results are shown in Appendix \ref{app:add_results}.} after $1 \times 10^6$ environment steps. Hopper Hop is a very challenging task even in the standard setting. Furthermore, the agents in these two tasks are tiny compared to the complex backgrounds, thus making it difficult for the model to distill task-relevant information. We note that the discrepancy between our results for CVRL and results reported in their paper \citep{ma2020contrastive} is due to the fact that while they used Imagenet \citep{russakovsky2015imagenet} for backgrounds in the original experiments, we use videos from kinetics dataset \citep{kay2017kinetics} instead. 

DBC fails to work on all six tasks, which is in contrast to their reported performance in \citep{zhang2020learning}. We note that, in their experiments, the authors used a single video for both training and testing. Their poor performance in our setting suggests that DBC is incapable of dealing with multiple distracting scenes, and also does not generalize well to unseen distractors.


Overall, the empirical results show that our proposed model is superior to all the baselines. In practice, it is desirable to have model which can work well for both standard (easy) setting and natural background (complex, distracting) setting. \RCPC satisfies this, since it performs on-par with Dreamer and significantly better than CVRL and DBC on standard control tasks, and in natural background setting which Dreamer fails, \RCPC works reasonably well and outperforms their counterparts CVRL and DBC.


\subsection{Ablation Analysis}

We conduct an ablation analysis to evaluate the importance of each CPC objective employed in \RCPC. To do so, we compare the original model with two variants, where we omit the temporal predictive coding or static predictive coding objective respectively (added to prevent map collapse; see \cref{sec:practical}). The relative performance of these models on three control tasks is shown in Figure \ref{fig:abaltion}. The original \RCPC achieves the best performance, while the variant with only static predictive coding (denoted SPC) fails across all three tasks. \RCPC without static predictive coding (Unstable-\RCPC) is unstable, since it faces the map collapsing problem, which leads to poor performance.

\subsection{Analysis in the Random Background Setting} \label{sec:rand_bg_analysis}
\begin{figure}[!h]
\centering
\subfigure[Reconstructions of a trajectory. $1^{\text{st}}$ row: ground-truth data, $2^{\text{nd}}$ row: Dreamer, $3^{\text{rd}}$ row: SPC, $4^{\text{th}}$ row: TPC]{
	\label{subfig:toy_viz}
	\includegraphics[width=0.7\linewidth]{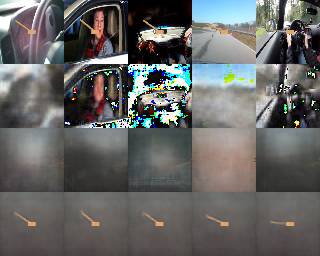}
} 
 
\subfigure[Performance of TPC versus Dreamer and SPC in the random background setting.]{
	\label{subfig:toy_results}
	\includegraphics[width=0.7\linewidth]{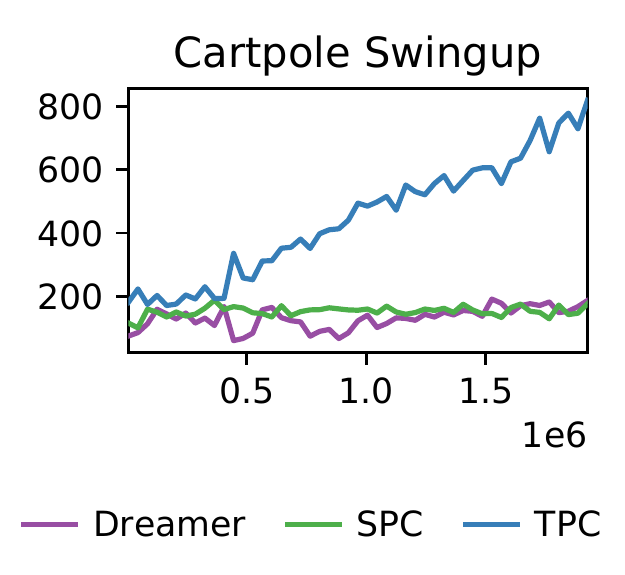}
} 
\caption{TPC and the baselines in the random background setting}
\label{fig:toy_exps}
\end{figure}

To demonstrate that temporal predictive coding is able to filter out temporally-unpredictive information, we conduct an experiment where the background is randomly chosen for each time step. In this setting, the only source of temporally-predictive information is the agent itself. Using the representations learned by Dreamer, \RCPC, and SPC, we then trained auxiliary decoders that try to reconstruct the original observations. \Cref{fig:toy_exps} shows that the representation learned by \RCPC is unable to reconstruct the background but can reconstruct the agent with high fidelity. In contrast, Dreamer's representation can reconstruct most of the image but not the agent itself, and SPC's reconstructions are largely uninterpretable. Furthermore, of the three models, only \RCPC successfully performs in the random background setting.

While the random background setting experiment aligns with our theoretical analysis and clearly demonstrates \RCPC's robustness to temporally-unpredictive nuisance information, it also exposes a particular limitation in our theoretical understanding of \RCPC's performance in the natural background setting, since natural videos are in fact a source of \emph{temporally-predictive} nuisance information. We provide two possible explanations for \RCPC's empirically strong performance in the natural background setting. First, the natural videos may contain considerable temporally-unpredictive information that \RCPC successfully dismisses. Second, the natural videos largely consist of hard-to-predict information that our recurrent neural network deems effectively unpredictable. We note that the latter potential explanation appears intimately related to the concept of ``usable information'' \cite{xu2020theory} whenever we instantiate a mutual information estimator with finite-computational power---and is worthy of further research scrutiny.


\section{Related Work}



\paragraph{Learning latent space for model-based RL via reconstruction } Latent world models can be learned by jointly training the encoder and the latent dynamics model with observation reconstruction loss. Learning Controllable Embedding (LCE) approach, including E2C \citep{watter2015embed}, RCE \citep{pmlr-v84-banijamali18a} and PCC \citep{Levine2020Prediction}, uses randomly collected data to pre-train a Markovian latent dynamic model that is specifically designed for locally-linear control, then run offline optimal control on top of the learned latent space. CARL \citep{cui2020control} extends these works for Soft Actor-Critic \citep{pmlr-v80-haarnoja18b} and also proposes an online version, in which they iteratively learn the model and a parameterized policy. World Models \citep{NIPS2018_7512} learn a recurrent latent dynamic model in a two-stage process to evolve their linear controllers in imagination. PlaNet \citep{pmlr-v97-hafner19a} jointly learns a recurrent state space model (RSSM) and plans in latent space using the cross entropy method, while Dreamer \citep{Hafner2020Dream} uses RSSM to iteratively learn the model and the policy by backpropagating through the dynamics. SOLAR \citep{pmlr-v97-zhang19m} models the dynamics as time-varying linear-Gaussian with quadratic costs and controls using guided policy search. However, training world models with reconstruction loss has several drawbacks: it requires a decoder as an auxiliary network for predicting images, and by reconstructing every single pixel, those methods are potentially vulnerable to task-irrelevant information such as an irrelevant background.

\paragraph{Learning latent space for model-based RL via constrastive learning} An alternative framework for learning latent world models is contrastive learning. Contrastive learning is a self-supervised learning technique that aims to learn representations by contrasting positive samples against negative samples without having to reconstruct images \citep{oord2018representation, chen2020simple}. Recently proposed contrastive learning methods have achieved significant successes in learning representations purely from unlabeled data, which include works by \citep{chen2020simple,chen2020big,bachman2019learning,henaff2019data,he2020momentum,tian2019contrastive}. \citet{pmlr-v97-poole19a} has also established a close connection between contrastive learning and mutual information maximization. In the context of RL, recent works have proposed to use this framework to accelerate RL from pixels in two distinct directions: 1) cast contrastive learning as an auxiliary representation learning task, and use model-free RL methods on top of the learned latent space \citep{oord2018representation,srinivas2020curl}; and 2) use contrastive learning in conjunction with learning a latent dynamics for planning in the latent space \citep{shu2020predictive,ding2020mutual,Hafner2020Dream}.

\paragraph{Learning to control from pixels with distractors}
To work well in real-world scenarios, a control-from-pixel method needs to be robust against distractors in the observation. Recent works have been proposed to address this problem, including CVRL \citep{ma2020contrastive} and DBC \citep{zhang2020learning}. Similar to \RCPC, CVRL employs a contrastive-based objective to learn the latent representation and latent dynamics that enable direct planning in latent space. However, they aim to maximize the mutual information between the current observation $o_t$ and its corresponding latent state $s_t$. This in theory still encourages the model to encode as much information in the observation as possible, including rich but task-irrelevant information such as the background. Our proposed \RCPC, in contrast, employs a \emph{temporal} predictive coding objective, which helps discourage the encoding of provably task-irrelevant information, as shown in \cref{sec:theory}. Taking a different approach, DBC was proposed to learn invariant representations for RL by forcing the latent space to preserve the on-policy bisimulation metric. This is in theory better than our contrastive objective, since it is capable of filtering out task-irrelevant but temporally-predictive information. However, while being theoretically sound, their implemented objective does not adhere to the true bisimulation metric. Specifically, they never solved the fixed point equation to compute the metric, but instead used a heuristic version of it, where they replaced the ground metric $d$ with Euclidean distance while comparing two distributions of the next state, with $d$ being the current bisimulation metric.


\section{Conclusion}

In this work, we propose a temporal predictive coding approach for planning in latent space. We show theoretically and experimentally that temporal predictive coding prioritizes the encoding of task-relevant components over temporally-unpredictive---and thus provably-irrelevant---information. This is critically different from reconstruction-based objectives as well as static contrastive learning objectives that only maximize the mutual information between the current observation and its latent code, which indiscriminately favor the encoding of both task-relevant and irrelevant information.
Our experiments show that temporally predictive coding outperforms state-of-the-art models in environments dominated by task-irrelevant information while remaining competitive on standard DMC tasks. 

\section*{Acknowledgements}

This research was supported by NSF (\#1651565, \#1522054, \#1733686), ONR (N00014-19-1-2145), AFOSR (FA9550-19-1-0024), FLI, and Amazon AWS.

\bibliography{icml2021}
\bibliographystyle{icml2021}
\clearpage
\appendix
\onecolumn
\icmltitle{Supplementary Materials to \sharedtitle}

\section{Hyper Parameters}
\subsection{Standard setting}

Dreamer, CVRL and \RCPC share the following hyperparameters:

\paragraph{Model components}
\begin{itemize}
    \item Latent state dimension: $30$
    \item Recurrent state dimension: $200$
    \item Activation function: ELU
    \item The action model outputs a tanh mean scaled by a factor of 5 and a softplus standard deviation for the Normal distribution that is then transformed using tanh \citep{pmlr-v80-haarnoja18b}
\end{itemize}

\paragraph{Learning updates}
\begin{itemize}
    \item Batch size: 50 for Dreamer and CVRL, 250 for \RCPC
    \item Trajectories length: 50
    \item Optimizer: Adam \citep{kingma2014adam} with learning rates $6 \times 10^{-4}$ for world model, $8 \times 10^{-5}$ for value and action.
    \item Gradient update rate: 100 gradient updates every 1000 environment steps.
    \item Gradient clipping norm: 100
    \item Imagination horizon: 15
    \item $\gamma = 0.99$ and $\lambda = 0.95$ for value estimation
\end{itemize}

\paragraph{Environment interaction}
\begin{itemize}
    \item The dataset is initialized with $S = 5$ episodes collected using random actions.
    \item We iterate between $100$ training steps and collecting $1$ episode by executing the predicted mode action with $\Normal(0, 0.3)$ exploration noise.
    \item Action repeat: 2
    \item Environment steps: $2 \times 10^6$
\end{itemize}

Additionally,
\begin{itemize}
    \item Dreamer and CVRL clip the KL below $3$ nats
    \item CVRL uses a bi-linear model for the critic function in the contrastive loss: $f_\theta(s_t, o_t) = \exp (z_t^T W_\theta s_t)$, where $z_t$ is an embedding vector for observation $o_t$ and $W_\theta$ is a learnable weight matrix parameterized by $\theta$.
    \item \RCPC has a fixed set of coefficient in the overall objective for all control tasks: $\lambda_1 = 1, \lambda_2 = 0.1, \lambda_3 = 1, \lambda_4 = 1$. We use a fixed Gaussian noise $\epsilon \sim \Normal(0, 0.2^2)$ to add to the future latent code when computing temporal CPC, as suggested in \citep{shu2020predictive}, and also use $0.2$ as the fixed variance in static CPC.
\end{itemize}

In \RCPC, we also use a target network for the value model and update this network every $100$ gradient steps. Note that we also tried to use target value network for Dreamer, but it does not improve the results, as suggested by their original paper \citep{Hafner2020Dream}.

\paragraph{Hyperparameters for DBC}
We use the same set of hyperparameters as reported in the paper \citep{zhang2020learning}
\begin{itemize}
    \item Replay buffer capacity: $1000000$
    \item Batch size: $128$
    \item Discount $\gamma$: $0.99$
    \item Optimizer: Adam
    \item Critic learning rate: $10 ^ {-5}$
    \item Critic target update frequency: $2$
    \item Critic Q-function soft-update rate $\tau_Q: 0.005$
    \item Critic encoder soft-update rate $\tau_\phi: 0.005$
    \item Actor learning rate: $10^{-5}$
    \item Actor update frequency: 2
    \item Actor log stddev bounds: $[-5,2]$
    \item Encoder learning rate: $10^{-5}$
    \item Decoder learning rate: $10^{-5}$
    \item Decoder weight decay: $10^{-7}$
    \item Temperature learning rate: $10^{-4}$
    \item Temperature Adam’s $\beta_1: 0.9$
    \item Init temperature: $0.1$
\end{itemize}

\paragraph{Hyperparameters search for \RCPC}
\RCPC has four hyperparameters that can be tuned: $\lambda_1$, $\lambda_2$, $\lambda_3$ and $\lambda_4$, which are coefficients for the TPC objective, consistency objective, SPC objective and reward prediction objective, respectively. Since $\lambda_1, \lambda_3$ and $\lambda_4$ do not conflict with each other, we fixed them to $1$ and only tuned $\lambda_2$ in our experiments. We performed grid search for $\lambda_2$ in range $\{0.05, 0.1, 0.2\}$\footnote{Larger values of $\lambda_2$ lead to representation collapse.} \emph{on the Cartpole Swingup task and then used the same set of hyperparameters for all the remaining tasks}.

\subsection{Natural background setting} \label{sec:a.2}

To further encourage the model to focus on task-relevant information from observations, we additionally tune the weight $\lambda_4$ of the reward loss in the training objective for both Dreamer and \RCPC. In each control task they share the same reward coefficient, which is specified in the table below. CVRL and DBC have the same hyperparameters as in the standard setting.
\begin{table}[ht]
\caption{Reward coefficients for different tasks in the natural backgrounds setting}
\label{table:reward_coefficient}
\begin{center}
\begin{tabular}{ll}
\textbf{Task} & \textbf{Reward coefficient} \\
\\ \hline \\
Cartpole Swingup, Cup Catch & 1000 \\
Cheetah Run, Walker Run, Pendulum Swingup, Hopper Hop      & 100
\end{tabular}
\end{center}
\end{table}

\newpage
\section{Proof of Lemma 1}\label{app:lemma}

Our goal is to show that, under the conditions in \cref{lem:relevance}, 
\begin{align}
    \eta(\pi^*) \ge \eta(\pi^*_\aux)
\end{align}
for any choice of auxiliary encoder $E'$. 

We start by denoting $s_t = E^*(o_t)$ and $s'_t = E'(o_t)$. Note that the performance of $\pi_\aux$ can be written as
\begin{align}
    \eta(\pi_\aux) &= \Expect_{(\pi_\aux,p)} r(o_\hor)\\
    &= \sum_{\tau_\aux} r(o_\hor) \prod_t p(o_t \giv o_\ltt, a_\ltt)p(s_t, s'_t\giv o_t)\pi_\aux(a_t \giv s_\leqt, s'_\leqt, a_\ltt),
\end{align}
where $\tau_\aux$ denotes the full trajectory of $(o, s, s', a)_\hor$ and $r(o_t)$ evaluates the reward at $o_t$ (for simplicity, we shall assume $p(r_t \giv s_t)$ is deterministic. Since $\kldiv{p(r_t \giv o_t)}{R^*(r_t \giv E^*(o_t))} = 0$, we can rewrite as
\begin{align}
    \eta(\pi_\aux) &= \sum_{\tau_\aux} R^*(s_\hor) \prod_t p(o_t \giv o_\ltt, a_\ltt)p(s_t, s'_t \giv o_t)\pi_\aux(a_t \giv s_\leqt, s'_\leqt, a_\ltt),
\end{align}
where, with a slight abuse of notation, we note that $R^*(E^*(o_t)) = r(o_t)$. We now further rewrite $\pi_\aux(a_t \giv s_\leqt, s'_\leqt, a_\ltt)$ as 
\begin{align}
    p(a_t \giv s_\leqt, s_\leqt, a_\ltt, \pi_\aux),
\end{align}
and subsequently collapse the expression of the performance as
\begin{align}
    \eta(\pi_\aux) 
    &= \sum_{(o, s, s', a)_\hor} R^*(s_\hor) p(o_\hor, s_\hor, s'_\hor, a_\hor \giv \pi_\aux)\\
    &= \sum_{(s, s', a)_\hor} R^*(s_\hor) p(s_\hor, s'_\hor, a_\hor \giv \pi_\aux),
\end{align}
where the last step arises from marginalization of $o_\hor$. Note by chain rule that $p(s_\hor, s'_\hor, a_\hor \giv \pi_\aux)$ becomes
\begin{align}
    \prod_t p(s_t \giv s_\ltt, s'_\ltt, a_\ltt, \pi_\aux) p(s'_t \giv s_\leqt, s'_\ltt, a_\ltt, \pi_\aux) p(a_t \giv s_\leqt, s'_\leqt, a_\ltt, \pi_\aux).
\end{align}
By analyzing the Markov blankets in $p(s_\hor, s'_\hor, a_\hor \giv \pi_\aux)$, we can simplify the above expression to
\begin{align}\label{eq:blanket}
    \prod_t p(s_t \giv s_\ltt, s'_\ltt, a_\ltt) p(s'_t \giv s_\leqt, s'_\ltt, a_\ltt) p(a_t \giv s_\leqt, s'_\leqt, a_\ltt, \pi_\aux).
\end{align}
Note that we omit the dependency on $\pi_\aux$ in the first two terms since, given only the history of past actions and observations, the next observation does not depend on our choice of policy but only on the environment dynamics. 

Since $E^*$ is optimal under the MI objective, we note that
\begin{align}\label{eq:equal}
    I(S_\ltt, S'_\ltt, A_\ltt \scolon S_t, S_t') =  I(S_\ltt, A_\ltt \scolon S_t).
\end{align}
\cref{eq:equal} implies that $s'_\ltt$ is independent of $s_t$ given $(s_\ltt, a_\ltt)$, and that $(s_\ltt, s'_\ltt, a_\ltt)$ is independent of $s'_t$ given $s_t$. This allow us to further simplify \cref{eq:blanket} to
\begin{align}
    \prod_t p(s_t \giv s_\ltt, a_\ltt) p(s'_t \giv s_t) \pi_\aux(a_t \giv s_\leqt, s'_\leqt, a_\ltt).
\end{align}
Thus, the performance expression equates to
\begin{align}\label{eq:augmented}
    \eta(\pi_\aux) &= \sum_{\tau_\aux} R^*(s_\hor) \prod_t p(s_t \giv s_\ltt, a_\ltt) p(s'_t \giv s_t) \pi_\aux(a_t \giv s_\leqt, s'_\leqt, a_\ltt).
\end{align}

Note by way of similar reasoning (up to and including \cref{eq:blanket}) that
\begin{align}\label{eq:standard}
    \eta(\pi) &= \sum_\tau R^*(s_\hor) \prod_t p(s_t \giv s_\ltt, a_\ltt) \pi(a_t \giv s_\leqt, a_\ltt).
\end{align}

By comparing \cref{eq:augmented} and \cref{eq:standard}, we see that $s'_\hor$ effectively serves as a source of noise that makes $\pi_\aux$ behave like a stochastic policy depending on the seed choice for $s'_\hor$. To take advantage of this, we introduce a reparameterization of $s'$ as $\epsilon$ such that
\begin{align}
    \eta(\pi_\aux) &= \sum_{\tau_\aux} R^*(s_\hor) \prod_t p(s_t \giv s_\ltt, a_\ltt) p(\eps_t) \pi_\aux(a_t \giv s_\leqt, \eps_\leqt, a_\ltt)\\
    &= \Expect_{p(\eps_\hor)}\sum_{(s, a)_\hor} R^*(s_\hor) \prod_t p(s_t \giv s_\ltt, a_\ltt) \pi_\aux(a_t \giv s_\leqt, \eps_\leqt, a_\ltt)\\
    &\le \max_{\eps_{1:T}}\sum_{(s, a)_\hor} R^*(s_\hor) \prod_t p(s_t \giv s_\ltt, a_\ltt) \pi_\aux(a_t \giv s_\leqt, \eps_\leqt, a_\ltt)\\
    &\le \max_{\pi} \eta(\pi),
\end{align}
where the last inequality comes from defining a policy
\begin{align}
    \pi' := \pi_\aux(a_t \giv o_\leqt, \eps^*_\leqt, a_\ltt)
\end{align}
and noting that the performance of $\pi'$ must be bounded by the performance of $\pi^*$. 
\hfill$\square$

\newpage
\section{Additional Results}\label{app:add_results}

\subsection{Comparision with CVRL after $2 \times 10^6$ environment steps}

\begin{figure}[ht]
    \centering
    \includegraphics[width=0.8\textwidth]{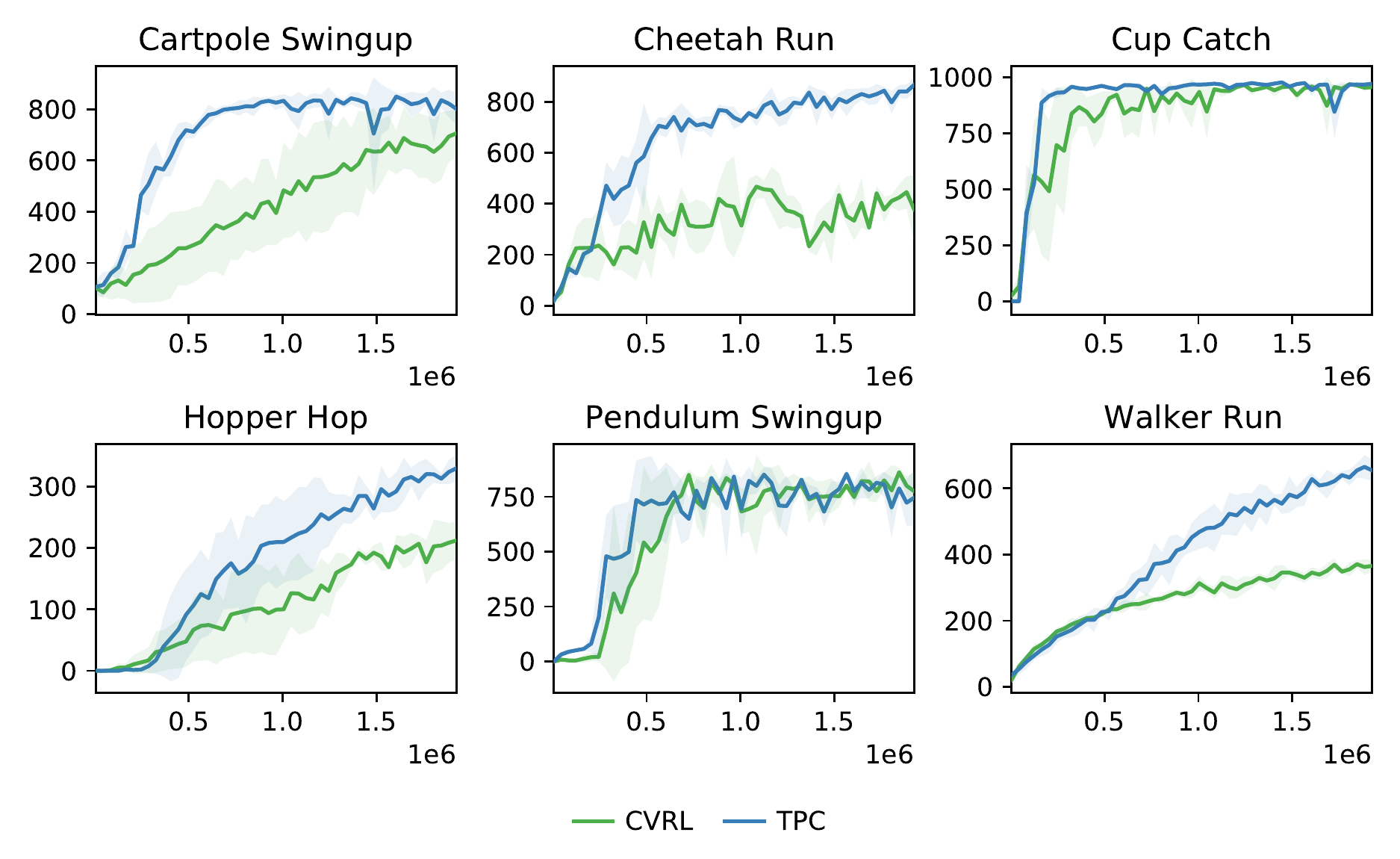}
    \includegraphics[width=0.8\textwidth]{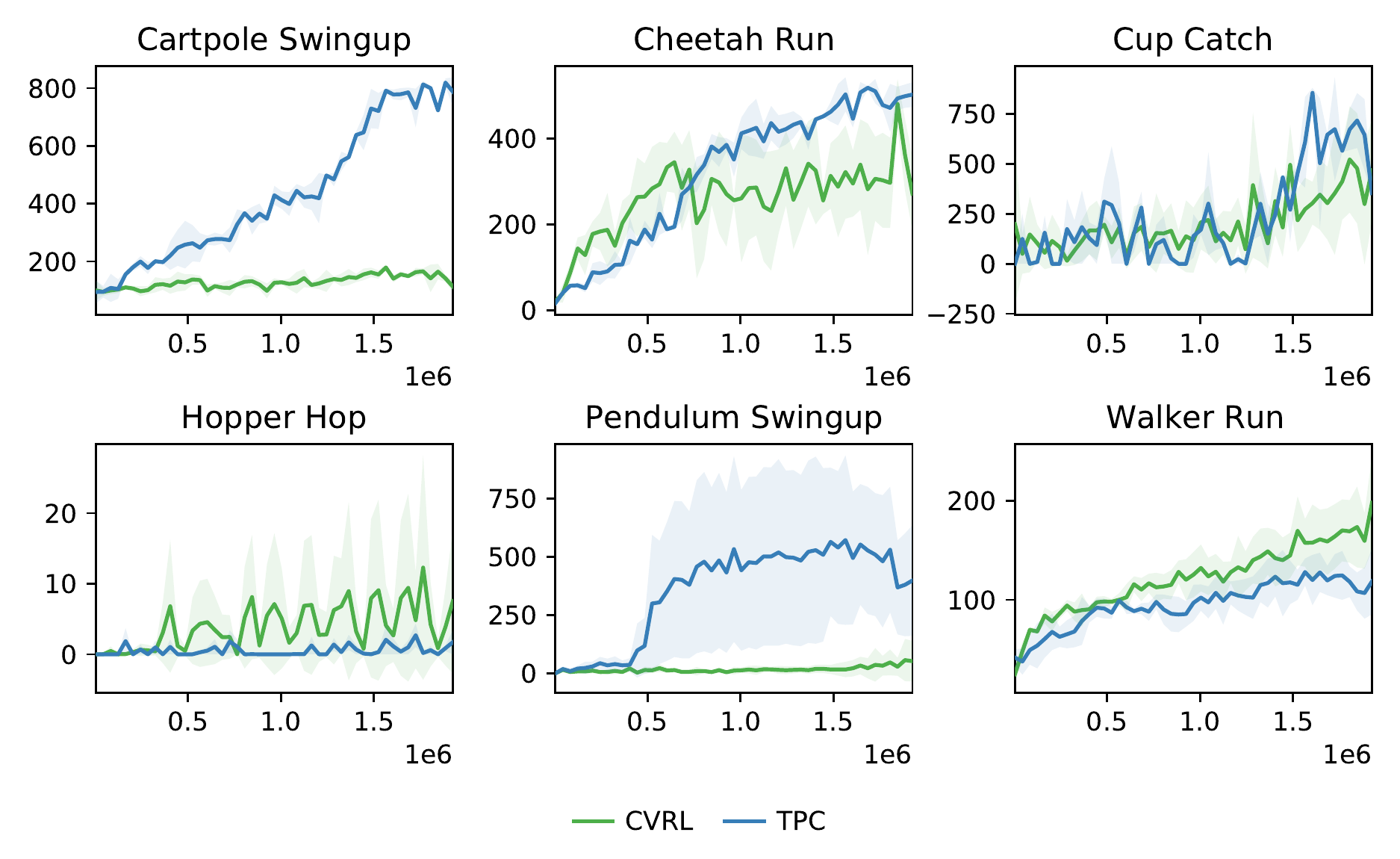}
    \caption{TPC vs CVRL after $2 \times 10^6$ in the standard setting (top) and background setting (bottom)}
    \label{fig:2m}
\end{figure}

In Figure \ref{fig:2m}, we compare the performance of \RCPC and CVRL after $2 \times 10^6$ environment steps in both the standard setting and the natural background setting. \RCPC learns faster and achieves much higher rewards compared to CVRL in the standard setting. In the natural control setting, \RCPC outperforms in $4$ out of $6$ tasks and is competitive in Walker Run. Both methods do not work on Hopper Hop.

\subsection{Comparison with CURL in the natural background setting}

\begin{figure}[h!]
    \centering
    \includegraphics[width=0.8\textwidth]{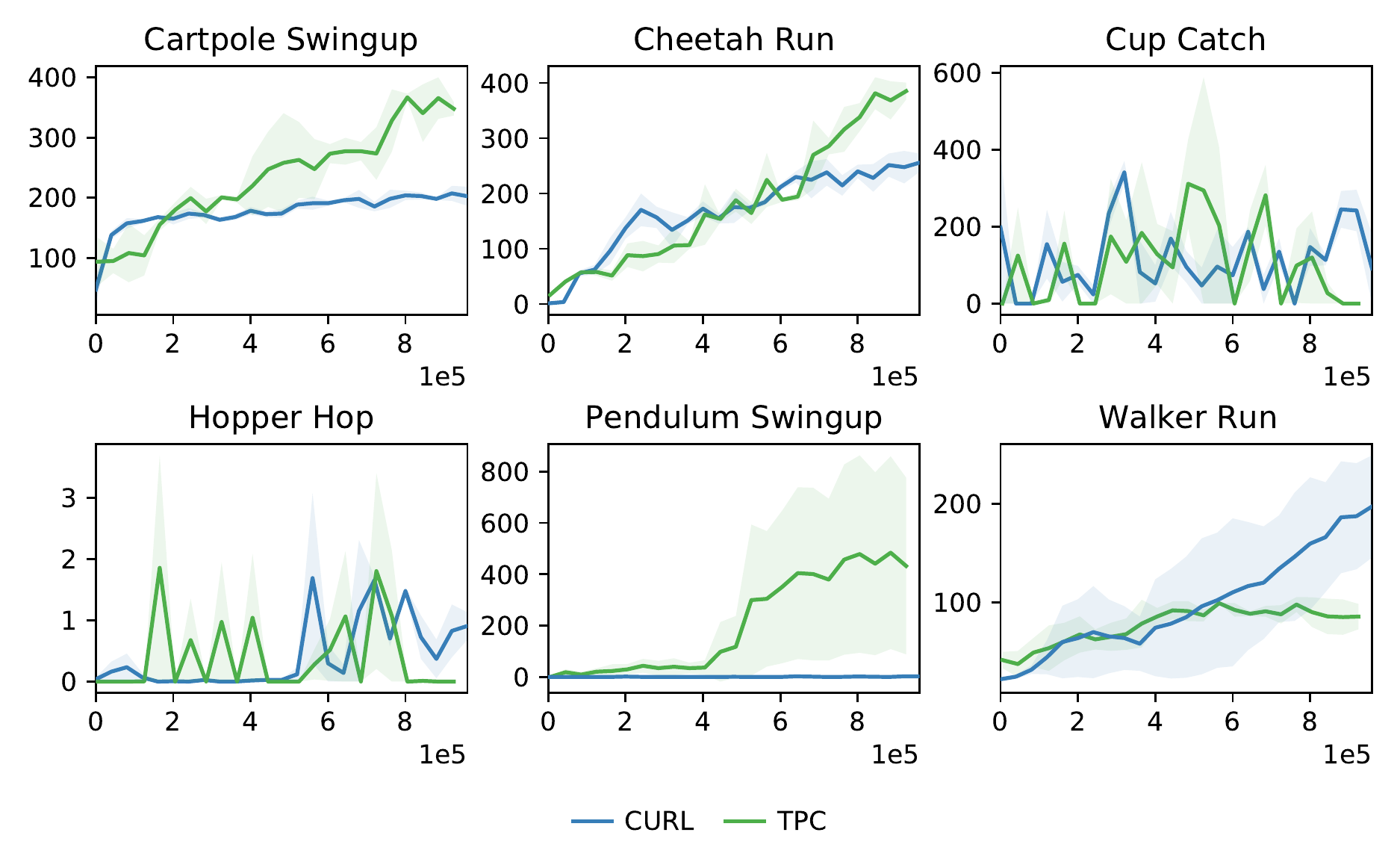}
    \caption{\RCPC versus CURL in the natural background setting. Each task is run with $3$ seeds.}
    \label{fig:compare_CURL}
\end{figure}

As shown in Figure \ref{fig:compare_CURL}, \RCPC outperforms CURL significantly on $3$ of $6$ tasks, while CURL performs better on Walker Run. On Hopper Hop and Cup Catch, both methods fail to make progress after $1$ million environment steps.

\subsection{Importance of dynamics smoothing}
\begin{figure}[h!]
    \centering
    \includegraphics[width=0.8\textwidth]{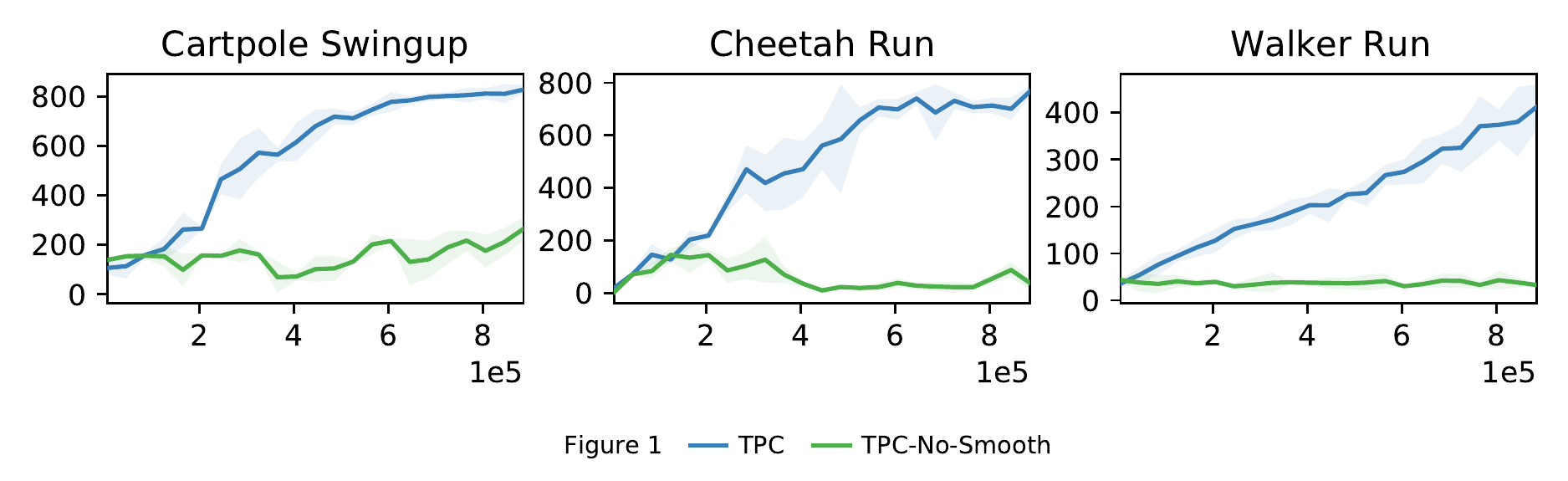}
    \caption{Performance of \RCPC in the standard setting without dynamics smoothing.}
    \label{fig:no_smooth}
\end{figure}

We run \RCPC in the standard setting without dynamics smoothing to investigate the empirical importance of this component. As shown in Figure \ref{fig:no_smooth}, \RCPC’s performance degrades significantly without dynamics smoothing. Without smoothing, the dynamics model cannot handle noisy rollouts during test-time planning, leading to poor performance. Dynamics smoothing prevents this by enabling test-time robustness against cascading error.

\subsection{Learning a separate reward model}
\begin{figure}[h!]
    \centering
    \includegraphics[width=0.3\textwidth]{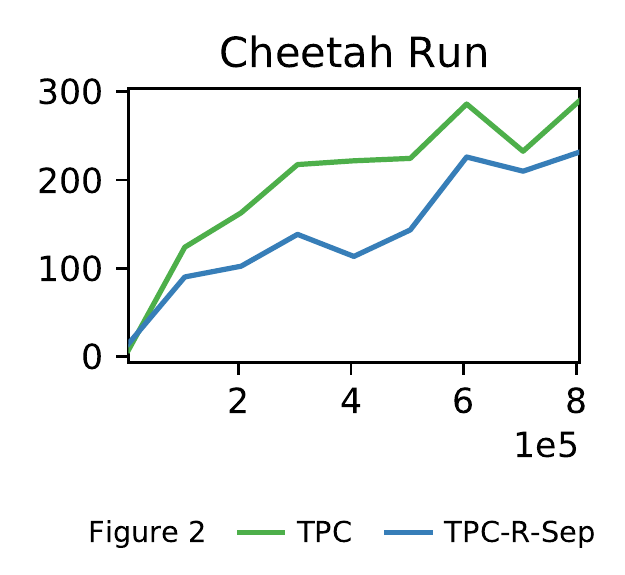}
    \caption{Performance of \RCPC in the random background setting with a separately trained reward model.}
    \label{fig:reward_separate}
\end{figure}

Joint learning of reward is crucial for all models in the natural background setting (see Appendix \ref{sec:a.2}). Since background information is also temporally-predictive, increasing the weight of reward loss encourages the model to focus more on the components that are important for reward learning. However, in the random background setting, since all temporally-predictive information is task-relevant, \RCPC \emph{uniquely} can learn the reward separately, as shown in Figure \ref{fig:reward_separate}.

\subsection{Reconstructions in the natural background setting}
\begin{figure}[ht]
    \centering
    \includegraphics[width=0.3\textwidth]{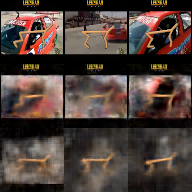}
    \caption{Observation reconstruction of \RCPC versus Dreamer in the natural background setting}
    \label{fig:natural_bg_recon}
\end{figure}

We conduct experiments to investigate what information the encoder in different models learns to encode during training in the natural background setting. To do that, we train auxiliary decoders that try to reconstruct the original observations from the representations learned by Dreamer and \RCPC. As shown in Figure \ref{fig:natural_bg_recon}, Dreamer ($2^{\text{nd}}$ row) tries to encode both the agent and the background. In contrast, \RCPC ($3^{\text{rd}}$ row) prioritizes encoding the agent, which is task-relevant, over the background.

\subsection{The simplistic motion background setting}
\begin{figure}[h]
    \centering
    \includegraphics[width=0.5\textwidth]{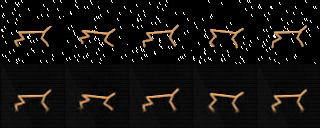}
    \caption{The top row shows a sample sequence of data, and the bottom row shows the reconstruction of \RCPC.}
    \label{fig:synthetic_viz}
\end{figure}
\begin{figure}[h]
    \centering
    \includegraphics[width=0.8\textwidth]{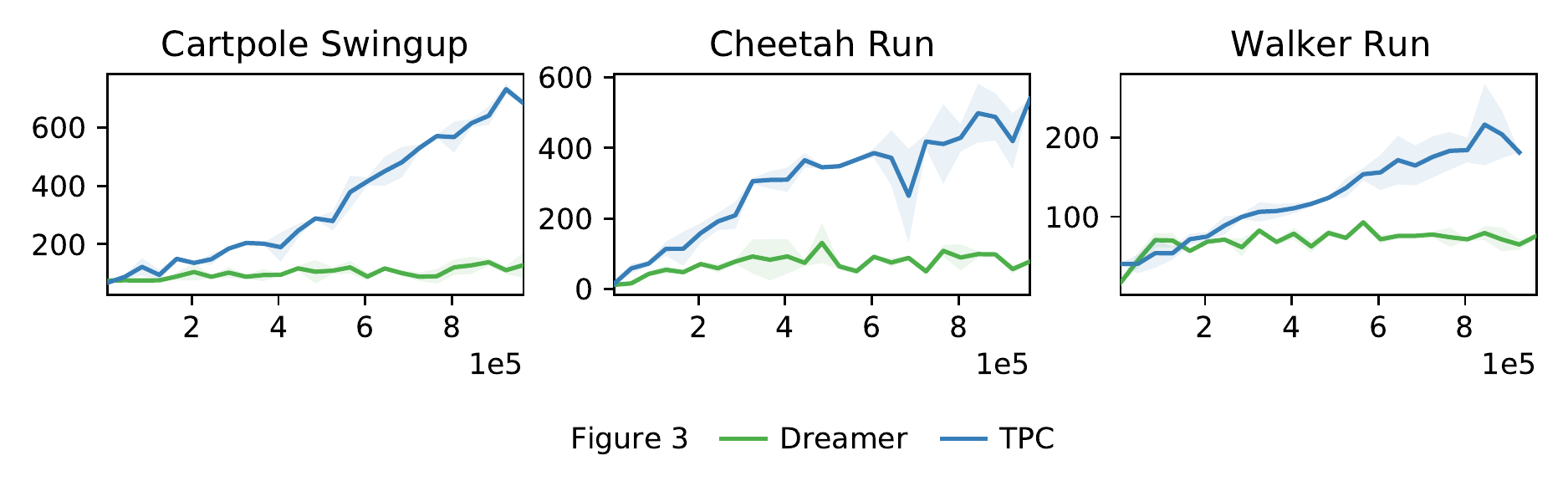}
    \caption{\RCPC versus Dreamer in the simplistic motion background setting}
    \label{fig:synthetic_exp}
\end{figure}

As discussed in Section \ref{sec:theory}, \RCPC can capture certain task-irrelevant information. However, \RCPC can choose to not encode the background \emph{whenever} encoding only the agent is sufficient to maximize the mutual information. In the experiments, we found that forcing the model to predict well the reward helps the encoder focus more on the agent, which can be done by increasing the weight of reward loss. Dreamer, in contrast, must encode as much information about the observation as possible to achieve a good reconstruction loss. To elaborate on this, we conducted an experiment where we replaced the natural background with a simplistic motion, easily predictable background, which is depicted in Figure \ref{fig:synthetic_viz}. Figure \ref{fig:synthetic_exp} shows that \RCPC works well in this setting, and outperforms Dreamer significantly.

\end{document}